\documentclass[letterpaper, 10 pt, conference]{ieeeconf}  % Comment this line out if you need a4paper

\IEEEoverridecommandlockouts                              % This command is only needed if 
\usepackage{graphicx} 
\usepackage{amsmath}
\usepackage{amssymb}  
\usepackage{amsfonts}
\usepackage{algorithmic}
\usepackage{array}
\usepackage{textcomp}
\usepackage{stfloats}
\usepackage{url}
\usepackage{verbatim}
\usepackage{graphicx}
\usepackage{caption}
\usepackage{subcaption}
\usepackage[ruled, vlined, linesnumbered]{algorithm2e}
\usepackage{cite}
\usepackage{amssymb,amsfonts,bm}
\usepackage{amsmath,tikz}
\usepackage{color}
\usepackage{mathtools}
\usepackage[hidelinks]{hyperref} 
\usepackage{rotating}
\usepackage{blkarray}
\usepackage{physics}
\usetikzlibrary{arrows}

\usepackage{amsthm}
\usepackage{comment}

\newcommand*{\ctrlin}{\vb{u}}

\newcommand*{\tor}{\tau}
\newcommand*{\mass}{m}
\newcommand*{\pos}{p}
\newcommand*{\posvec}{\vb{p}}

\newcommand*{\thrust}{T}
\newcommand*{\for}{f}

\newcommand*{\gbf}[1]{\boldsymbol{#1}}
\newcommand*{\distvec}{\vb{d}}

\newcommand*{\roll}{\phi}
\newcommand*{\pitch}{\theta}
\newcommand*{\yaw}{\gamma}
\newcommand*{\angvel}{\omega}
\newcommand*{\angvelvec}{\gbf{\omega}}
\newcommand*{\rotmatrix}{\vb{R}}
\newcommand*{\angvelwheel}{\angvel_w}
\newcommand*{\forcevec}{\vb{\for}}
\newcommand*{\torvec}{\gbf{\tor}}
\newcommand*{\screw}[1]{[#1]_\times}

\newcommand*{\radius}{r}

\newcommand*{\moi}{\vb{J}}

\newcommand*{\moiscaler}{J}
\newcommand*{\dist}{l}

\newcommand*{\angvelvecdesir}{\angvelvec_d}
\newcommand*{\velcontrol}{\torvec}

\newcommand*{\airk}{\vb{K}^\omega}
\newcommand*{\airka}{\vb{K}_p}
\newcommand*{\airkp}{\airk_p}
\newcommand*{\airki}{\airk_i}
\newcommand*{\airkd}{\airk_d}

\renewcommand{\vb}[1]{\boldsymbol{#1}}
\newcommand*{\scale}{\alpha}

\newcommand*{\degree}{^{\circ}}

\newcommand*{\kthrust}{C_{t}}
\newcommand*{\kdrag}{C_{d}}
\newcommand*{\motorvelo}{v}
\newcommand*{\motorvelovect}{\vb{v}}
\newcommand*{\ld}{l}
\newcommand*{\vmin}{v_{min}}
\newcommand*{\vmax}{v_{max}}
\title{\LARGE \bf
AirCrab: A Hybrid Aerial-Ground Manipulator with An Active Wheel
}

\author{Muqing Cao*, Jiayan Zhao*, Xinhang Xu, and~Lihua~Xie$^{1}$,~\IEEEmembership{Fellow,~IEEE}% 
\thanks{*Equal Contribution}% <-this % stops a space
\thanks{All authors are with the School of Electrical and Electronic Engineering,
        Nanyang Technological University, 50 Nanyang Avenue, Singapore.
        }
\thanks{$^{1}$ Corresponding author}%%
}

\begin{document}

\maketitle
\thispagestyle{empty}
\pagestyle{empty}

%%%%%%%%%%%%%%%%%%%%%%%%%%%%%%%%%%%%%%%%%%%%%%%%%%%%%%%%%%%%%%%%%%%%%%%%%%%%%%%%
\begin{abstract}

Inspired by the behavior of birds, we present AirCrab, a hybrid aerial ground manipulator (HAGM) with a single active wheel and a 3-degree of freedom (3-DoF) manipulator. 
AirCrab leverages a single point of contact with the ground to reduce position drift and improve manipulation accuracy.
The single active wheel enables locomotion on narrow surfaces without adding significant weight to the robot.
To realize accurate attitude maintenance using propellers on the ground, 
we design a control allocation method for AirCrab that prioritizes attitude control and dynamically adjusts the thrust input to reduce energy consumption.
Experiments verify the effectiveness of the proposed control method and the gain in manipulation accuracy with ground contact.
A series of operations to complete the letters `NTU' demonstrates the capability of the robot to perform challenging hybrid aerial-ground manipulation missions.
%To reduce power consumption, we propose an attitude control allocation strategy on the ground which saves energy and gets higher accuracy when compared to Ardupilot attitude controller. By the single contact point, AirCrab can work in rough and narrow terrain and achieve operation accuracy comparable to that of a static manipulator cost-effectively. A challenging mission demonstrates the capability of grasping, taking off with payload, landing on a table, and building the characters 'NTU'.

\end{abstract}

%%%%%%%%%%%%%%%%%%%%%%%%%%%%%%%%%%%%%%%%%%%%%%%%%%%%%%%%%%%%%%%%%%%%%%%%%%%%%%%%
\section*{Supplementary Material}
A video summarizing the approach and experiments is available at
https://youtu.be/Q1n-IiIt400. 

\section{INTRODUCTION}
In recent years, new aerial robots have emerged that perform tasks requiring physical interaction with the surroundings. These include aerial robots equipped with (1) robot arms to pick and place objects \cite{ollero2021past}, (2) contact-based sensing devices for data collection \cite{7353623}, and (3) ground-driving wheels for hybrid air-ground locomotion \cite{cao2023doublebee}. 
These emerging aerial robots open up many application scenarios, such as item manipulation and transportation \cite{appius2022raptor}, contact-based façade and bridge inspection \cite{7353623}, and long-duration exploration of unknown regions. 
However, the need for high robustness and accuracy in the interaction poses a significant challenge for aerial robots, 
whose control and localization accuracy are affected by complex aerodynamic effects, wind disturbances, and the limitation of the onboard sensors and computational resources \cite{Suarez2020benchmark}.
%which are inherently subject to complex aerodynamic effects and wind disturbances, especially when operating near objects or surfaces.
%The limitations of the localization accuracy using onboard sensors prevent 
The highly coupled position and attitude of the traditional multi-rotor robots also make it challenging to perform accurate contact.

Various solutions have been proposed in the context of contact-based inspection.
In \cite{rashad2020towards, Bodie2021active}, fully actuated UAVs are designed to achieve decoupled position and attitude control for precise sensing using fixed-mounted contact probes. 
In a different research direction, multi-rotor robots are equipped with several top-mounted wheels for inspection of ceiling or bridge floor
\cite{ladig2016high,wang2020measurement}. 
During the inspection, thrust from rotors pushes the robot against the surface while the wheels in contact with the top surface realize the locomotion.
Such designs reduce the degree of freedom of the robots and effectively alleviate the position drift and attitude oscillation during the inspection.
However, adding numerous active and passive wheels increases the weight and the power consumption.
%Instead of adding more actuators for flight, contacted-based methods\cite{ladig2016high,wang2020measurement} only need the vehicle to contact with the top surface and use omnidirectional wheels or a combination of active wheels and passive wheels on the top to move on the top surface. 
%So that the DoF of the vehicle will be constrained by the top surface and the underactuation degrees will be decreased so that it is easier to control the system than freely flying in the air. 
%These designs do increase the operation accuracy, but also the extra actuators can not be used for saving a lot of energy in their design.

In the context of aerial manipulation (AM), a new class of robots called hybrid aerial-ground manipulators (HAGM) has been proposed \cite{ollero2021past}.
Incorporating design principles from hybrid aerial-ground vehicles (HAGVs) \cite{cao2023doublebee,pan2023skywalker,zheng2023roller}, HAGMs are equipped with ground-driving mechanisms to reduce energy consumption and improve manipulation accuracy by exploiting surface contact.
In \cite{lopez2020mhyro}, a hexarotor UAV is equipped with a crawler and roller for locomotion on pipes while inspecting the pipes by a 3-DoF arm, but the heavy driving mechanisms compromise the aerial efficiency. 
In \cite{saeed2018modeling, salman2020design}, a HAGM design is proposed with two active wheels and two passive wheels but no physical robot is built.
Chat-PM\cite{ding2024chat} combines a dual-passive-wheel HAGV \cite{zhang2022autonomous} with a lightweight 4-DoF arm to achieve the state-of-the-art operation accuracy. 
However, the use of passive wheels means that attitude and position are still highly coupled in ground motion, reducing its flexibility in manipulation tasks. 
In general, the existing HAGMs employ a multi-wheel design requiring multiple contact points, hindering applications on narrow surfaces.

\begin{figure}[]
      \centering      \includegraphics[width=\linewidth]{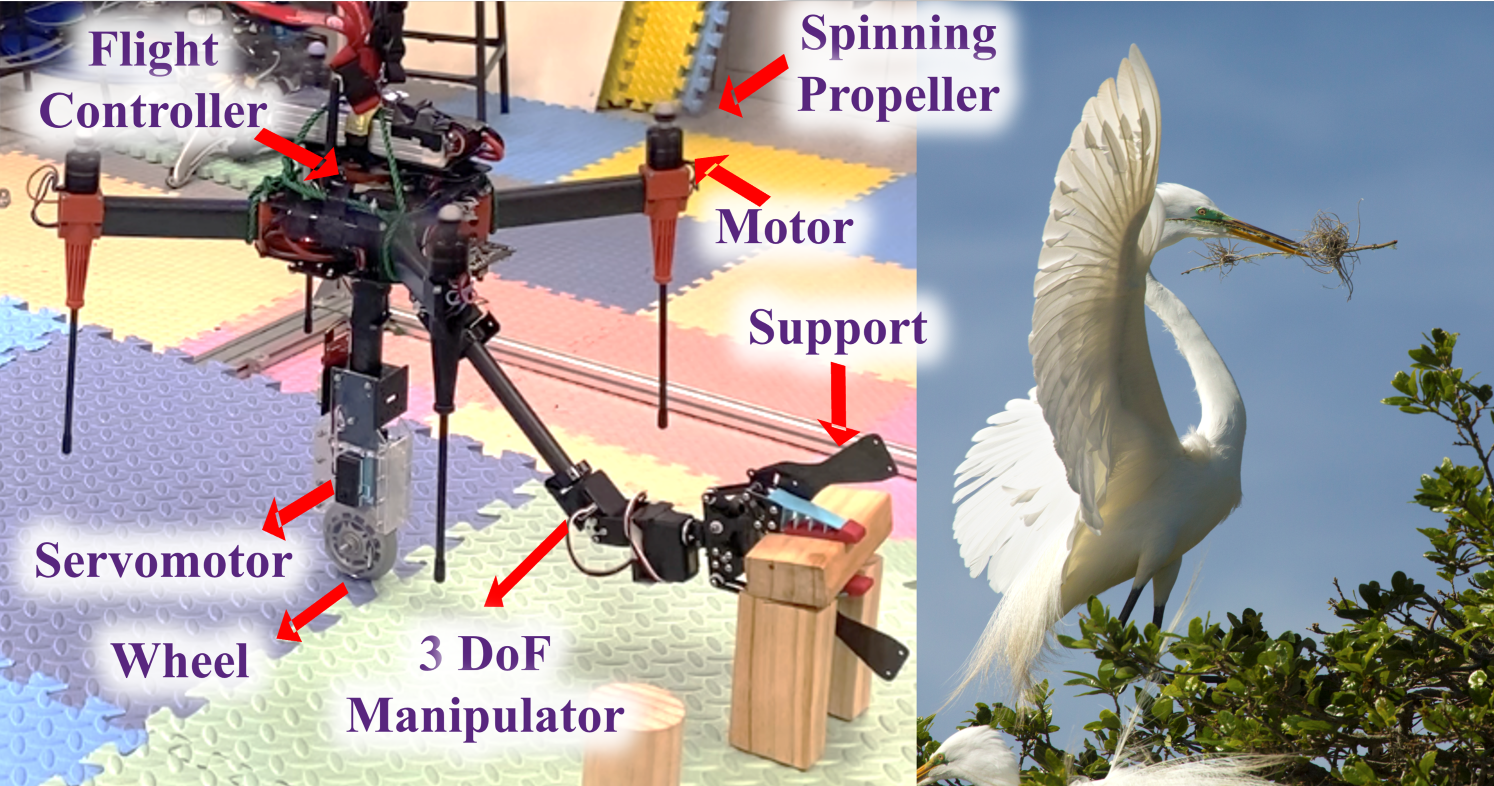}    
      \caption{\footnotesize The AirCrab mimics the nesting behavior of egrets, standing on a single active wheel, maintaining attitude with propellers, and performing precise operation. }
      \label{fig: face}
\end{figure}

In this work, we propose AirCrab, a design of HAGM with only one active wheel.
Such a design is inspired by the behavior of egrets who can stand on a single leg to provide an anchor for precise manipulation and flap wings to maintain balance, as shown in Fig. \ref{fig: face}.
The lightweight active wheel provides a single contact point on narrow terrain and does not add much redundant weight in the aerial operation.
The single contact point facilitates rotation around the robot's vertical axis, adding a degree of freedom for manipulation.
Furthermore, the active wheel enables decoupled position and attitude control, where the propellers are used to maintain attitude while the translational motion is realized by wheel rotation.
To reduce energy consumption and improve control performance on the ground, we design a control allocation approach for the propellers that prioritizes attitude maintenance and dynamically selects the minimum necessary propeller thrust. 
The main contributions of this paper are summarized as:
\begin{itemize}
    \item A design of a HAGM with an active wheel and a 3-DoF manipulator is proposed and built; 
    \item Detailed dynamics modeling and controller design are carried out, and a control allocation strategy is proposed to prioritize attitude control and reduce propeller thrust;
    \item Experimental results verify the high accuracy and energy efficiency of the proposed control allocation strategy and the trajectory tracking accuracy of the end effector in different modes;
    \item Challenging missions showcase the capability of AirCrab in grasping an object on the ground, taking off with payload, landing on a table, and placing the object to complete the characters `NTU' (Fig. \ref{fig: NTU}). 
\end{itemize}

The rest of the paper is organized as follows:
 Section \ref{sec: Mech} describes the mechanical design of AirCrab. The detailed dynamics model is shown in the section \ref{sec: dynamics}. Section \ref{sec: controller} explains the attitude controller design and presents the proposed control allocation strategy. 
 Section \ref{sec: exp} details the experimental verification in terms of attitude control, power consumption, manipulator tracking performance, and the execution of complex hybrid aerial-ground missions. 
 Finally, section \ref{sec: conclusions} concludes the paper.

\section{Mechanical Design}\label{sec: Mech}
We construct a prototype based on a general quadrotor design with a diagonal length of $ 450$ mm, equipped with four brushless HS$2216$ $920$KV motors and $10$-inch propellers.
The onboard Cube Orange flight controller runs a customized Ardupilot firmware for attitude and position control. 
The total weight of the robot including the wheel, the manipulator, the quadcopter and a 4S 5200 mAh battery is 2.655 kg.

% \begin{figure}[]
%       \centering      \includegraphics[width=\linewidth]{figure/mech_struct_compact.png}    
%       \caption{Detailed mechanical design of flying crab}
%       \label{fig: mech_strucure}
% \end{figure}

\begin{figure}[]
      \centering      \includegraphics[width=\linewidth]{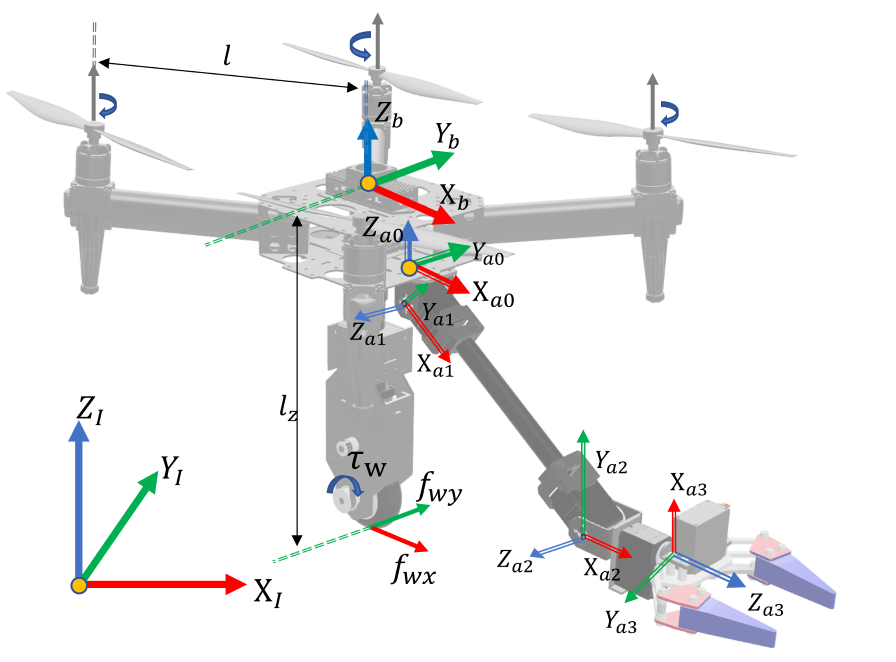}    
      \caption{\footnotesize Coordinate frames for AirCrab. The inertial frame is labeled with a subscript $I$ and the body frame is labeled with a subscript $b$.}
      \label{fig: axes}
\end{figure}

A servo-driven rubber wheel is connected to the bottom plate of the quadrotor chassis through a carbon fiber tube, as shown in Fig  \ref{fig: face}.
A lightweight servomotor STS3020 with 24.3 kg·cm maximum torque capability is chosen. 
To ensure the wheel lies directly below the geometric center of the quadrotor, we use a belt and two pulleys to drive the wheel instead of connecting the servo directly to the wheel. 
The total weight of the wheel, the servomotor, and the mechanical connections is $305$ g.

The $3$-DoF manipulator arm utilizes three rotational joints to offer 2 DoF in position and 1 DoF in rotation for the end effector.
The rotation axes for the joints are shown in Figure \ref{fig: axes}.
A gripper is mounted as the end effector for object pick and place.
Each joint and gripper is driven by a servomotor and connected to an ESP32-based driver board for control.
Two extra carbon fiber plates are mounted on the end effector to provide two widely spaced support points, forming a triangular support for AirCrab when powered off.

\section{Dynamics Modelling}\label{sec: dynamics}
The translational and rotational dynamics of AirCrab can be expressed as: 
\begin{align}
&\mass\ddot{\posvec}=\vb{\for}_g+\rotmatrix(\vb{\for}+\vb{\for}_{w} + \vb{\for}_{d}),\label{eq: forcebody}\\
&\moi\dot{\angvelvec}+\screw{\angvelvec}\moi\angvelvec = \gbf{\tor} - \gbf{\tor}_w + \gbf{\tor}_d + \screw{\distvec_w}\vb{\for}_w + \screw{\distvec_g}\vb{\for}_g,\label{eq: torquebody}
\end{align}
where $\posvec=[\pos_x,\pos_y,\pos_z]^\top$ denotes the position of the robot in the inertial frame; 
$\gbf{\angvel}\in\mathbb{R}^3$ denotes the angular velocity in the body frame;
$\rotmatrix$ denotes the rotation matrix from the robot body frame to the inertial frame;
$\mass\in\mathbb{R}$ and $\moi\in\mathbb{R}^{3\times3}$ denote the mass and the moment of inertial about the body frame of the robot; 
$\vb{\for}=[0, 0, \thrust]^\top$ and $\gbf{\tor}\in\mathbb{R}^3$ denote the collective thrust and torque generated by the rotors in the body frame;
$\vb{\for}_w$ denotes the force acting on the wheel due to contact with the ground and $\gbf{\tor}_w=[0, \tor_w,0]^\top$ represents the torque exerted by the servomotor on the wheel;
$\vb{\for}_d$ and $\gbf{\tor}_d$ represent the force and torque arising from disturbances due to the movements of the manipulator and aerodynamic effects;
$\distvec_w$ and $\distvec_g$ denote the vector from the robot's center of gravity (CG) and ground contact point to the origin of its body frame, expressed in body and inertial frame, respectively.
$\screw{\cdot}$ denotes the screw-symmetric cross product of a vector.

When the robot is in the aerial mode, the wheel is not in contact with the ground, $\vb{\for}_w$ and $\gbf{\tor}_w$ are equal to zero. 
Hence the dynamics model \eqref{eq: forcebody} and \eqref{eq: torquebody} becomes identical to those of a traditional multi-rotor UAV \cite{Mellinger2011minimum}.

When the robot is in contact with the ground, $\vb{\for}_w$ and $\gbf{\tor}_w$ are nonnegligible.
We specifically consider the case where the attitude of the robot is maintained near the level condition, i.e., roll and pitch angles $\roll,\pitch\approx0$.
In practice, this can be achieved by controlling the rotor speed, as will be shown in Section \ref{sec: exp}.
In this case, $\rotmatrix$ can be approximated as a yaw-only rotation around the inertial $z$-axis, $\rotmatrix_z(\yaw)$.
As a result, $\rotmatrix\forcevec\approx[0, 0, \thrust]^\top$, the effect of rotor thrust on the x and y movement of the robot is negligible.
The 2D translation of the robot on the ground can be approximated as
\begin{align}
    \mass\begin{bmatrix}\ddot{\pos}_{x} \\ \ddot{\pos}_{y} \end{bmatrix}\approx\begin{bmatrix}
\cos{\yaw} & -\sin\yaw \\
\sin\yaw & \cos\yaw 
\end{bmatrix}\left(\begin{bmatrix}\for_{wx} \\ \for_{wy} \end{bmatrix}
+\begin{bmatrix}\for_{dx} \\ \for_{dy} \end{bmatrix}\right),
\end{align}
where $\for_{wx},\for_{wy},\for_{dx},\for_{dy}$ are the $x$ and $y$ components of $\forcevec_w$ and $\forcevec_d$, respectively.
$\for_{wx}$ and $\for_{wy}$ indicate the frictional force at the ground contact point. 
When the wheel is stationary, small disturbances acting on the robot are compensated by the static frictional forces, reducing the translational drifts due to disturbances.

When the robot maintains a level attitude, the ground contact point lies close to the $z$-axis of the body frame, i.e., $\distvec_w\approx[0, 0, -\dist_{z}]$ for a $\dist_z>0$. 
Referring to Equation \eqref{eq: torquebody}, the effect of $\forcevec_w$ and $\forcevec_g$ on the yaw rotation can be neglected.
Furthermore, given that the torque input $\torvec_w$ does not apply on the body $z$-axis (Figure \ref{fig: axes}), the yaw rotation in ground mode is similar to that in aerial mode,  dominated by the torques from rotors and disturbances.
Clearly, the influence of ground contact on the roll and pitch dynamics is more significant than on the yaw dynamics.
% can be simplified from Equation \eqref{eq: torquebody} as 
% \begin{align}
%     &\moiscaler_z\dot{\angvel}_z+\angvel_z\moiscaler_z\angvel_z = \tor_z + \tor_{dz} ,\label{eq: torquez}
% \end{align}
% where $\tor_z$ and $\tor_{dz}$ are the $z$-component of $\torvec_w$ and $\torvec_d$, respectively, $\moiscaler_z$ is the third diagonal element of $\moi$ and the off-diagonal elements are assumed to be small. 
%Equation \eqref{eq: torquez} indicates that the yaw rotational dynamics on the ground are similar to that in the aerial mode, controlled mainly by the rotors. 
Finally, the wheel dynamics and kinematics can be expressed as:
\begin{align}
    &\moiscaler_w\dot{\angvel}_w=\tor_w-\radius\for_{wx},\\
    &\angvelwheel\radius=\dot{\pos}_{wx},
\end{align}
where $\angvelwheel$ and $\radius$ are the rotation rate and the radius of the wheel, $\pos_{wx}$ is the position of the wheel in the body $x$-axis, $\moiscaler_w$ indicates the moment of inertia of the wheel about its rotation axis.

\section{Controller Design}\label{sec: controller}
\subsection{Attitude Control}
The control inputs for AirCrab are the rotor thrust $\thrust$, the rotor torques $\torvec$, and the servomotor torque $\tor_w$.
In the aerial mode, desired thrust $\thrust$ and the desired attitude $\rotmatrix_d$ are computed from the high-level position controller similar to \cite{Mellinger2011minimum}.
In the ground mode, the desired attitude is set to a level condition $\rotmatrix_d=\rotmatrix_z(\yaw_d)$, where $\yaw_d$ is the desired yaw angle obtained from pilot input or high-level planner.
The thrust in ground mode is dynamically adjusted in the control allocation as will be explained in Section \ref{subsec: allocation}.
Given the desired attitude, the input torque is computed using a cascade PID controller:
\begin{align}
&\angvelvecdesir  =\airka\frac{1}{2}(\rotmatrix^\top_d\rotmatrix-\rotmatrix^\top\rotmatrix_d)^\vee,\\
&\angvelvec_e  =\angvelvecdesir-\angvelvec,\\
&\velcontrol  =\airkp \angvelvec_e +\airki \int \angvelvec_e+\airkd  \dot{\angvelvec}_e + \torvec_c,
\end{align}
where $\angvelvecdesir$ and $\angvelvec_e$ are the desired angular velocity and the angular velocity error, $\airka, \airkp, \airki, \airkd$ are the diagonal matrices consisting of the PID gains, $^\vee$ is the mapping from $so(3)$ to $\mathbb{R}^3$. 
$\torvec_c\in\mathbb{R}^3$ is added to compensate for the off-center CG.
For simplicity, we set $\torvec_c$ to be constant in this work, however, 
techniques from existing literature \cite{Cao2023eso} can be employed to compensate for changing disturbances.
The torque to drive the wheel is computed from the desired rotation rate $\angvel_{wd}$:
\begin{align}
    \tor_w=K\angvel_{wd},
\end{align}
where $K$ is a proportion gain. $\angvel_{wd}$ is obtained from the high-level planner or the pilot input.

\subsection{Control Allocation in Ground Mode} \label{subsec: allocation}
%In this part, a control allocation methods used in ground mode is proposed, which prioritize tilt control over torsion control, and minimize the thrust generated in ground mode to realize energy saving.
Among the computed control inputs, $\tor_w$ can be converted directly to a signal and sent to the servomotor. 
On the other hand, $\thrust$ and $\torvec$ have to be collectively realized by four propellers due to the coupling between row, pitch, yaw, and thrust. 
For a frame configuration as AirCrab (commonly called an X frame), there exists a linear mapping from quadratic motor speeds of the four motors $\motorvelovect=[\motorvelo_1^2, \motorvelo_2^2, \motorvelo_3^2, \motorvelo^2_4]^\top$ to  control inputs $\ctrlin=[\thrust,\tor^\top]^{\top}\in \mathbb{R}^{4}$, more specifically, $\ctrlin=\vb{M}\motorvelovect$ can be written as 
% The forces and torques generated by the propulsion groups can be written as:
% \begin{align}
%     &\for_{i}=\kthrust\motorvelo^{2},\label{eq: thrust}\\
%     &\tor_{i}=\kdrag\motorvelo^{2},\label{eq: drag}
% \end{align}
\begin{align}
    \begin{bmatrix}
     \thrust \\ \tor_{x} \\ \tor_{y}  \\ \tor_{z}
    \end{bmatrix} = \begin{bmatrix}
        \kthrust & \kthrust & \kthrust & \kthrust \\
        -\frac{\sqrt{2}}{2}\kthrust \ld & -\frac{\sqrt{2}}{2}\kthrust \ld & \frac{\sqrt{2}}{2}\kthrust \ld & \frac{\sqrt{2}}{2}\kthrust \ld \\
        -\frac{\sqrt{2}}{2}\kthrust \ld &\frac{\sqrt{2}}{2}\kthrust \ld & -\frac{\sqrt{2}}{2}\kthrust \ld & \frac{\sqrt{2}}{2}\kthrust \ld \\
        -\kdrag & \kdrag & \kdrag & -\kdrag
    \end{bmatrix}
    \begin{bmatrix}
        v_{1}^{2} \\ v_{2}^{2} \\ v_{3}^{2} \\ v_{4}^{2} 
    \end{bmatrix}\label{eq: mapping}
\end{align}
where $\kthrust>0$ and $\kdrag>0$ are the thrust coefficient and the drag coefficient, respectively, and $\ld$ is the distance between propellers and  geometry center of frame. 
%$i=1,2,3,4$. 
%From frame configuration and \eqref{eq: thrust},\eqref{eq: drag}, the linear mapping from quadratic motor speed $\motorvelovect\in\mathbb{R}^{4}$ to  control inputs $\ctrlin=[\thrust,\tor]^{T}\in \mathbb{R}^{4}$ characterized by matrix $\vb{M}\in \mathbb{R}^{4\times4}$ is defined as:

%where $\ld$ is the distance between propellers and  geometry center of frame. 
It may seem straightforward to apply the inverse mapping $\motorvelovect = \vb{M}^{-1}\ctrlin $ to attain the motor speeds; however, the results may exceed the mechanical limits of the motors, as each motor is subject to velocity constraints, $0\leqslant\vmin \leqslant v_{i} \leqslant \vmax$. 
Therefore, a control allocation problem is solved to find a modified control input $\hat{\ctrlin}=\vb{g}(\ctrlin)$ such that $\motorvelovect = \vb{M}^{-1}\hat{\ctrlin} \in\mathbb{V}$, where $\mathbb{V} = \left\{ \motorvelovect \in \mathbb{R}^{4} |\vmin^2 \leqslant v^2_{i} \leqslant \vmax^2 \right\}  \label{eq: v}$.
Common control allocation approaches are weighted least square (WLS) \cite{zhou2010reconfigurable}, direct control allocation (DCA)\cite{durham1993constrained} and partial control allocation (PCA) \cite{monteiro2016optimal}.
%A common solution is the direct control allocation (DCA) method \cite{durham1993constrained}, which computes a scaled input $\hat{\ctrlin}=\scale\ctrlin$, $\scale\in(0,1]$ to satisfy the constraints.
Our approach is based on the PCA which groups the inputs into different priorities and preserves the inputs with the higher priority as much as possible.
We choose to classify $\ctrlin$ into the tilt input ($\tor_{x},\tor_{y}$), yaw input ($\tor_z$), and the thrust input ($\thrust$), in the order of decreasing priority.
The motivation is that more control effort should be allocated to the roll and pitch, which faces larger disturbances from the ground contact as analyzed in Section \ref{sec: dynamics}.
Furthermore, with ground support, the thrust from the motors is not necessary to maintain the position and should be minimized to reduce energy consumption.

The control allocation strategy is presented in Algorithm \ref{alg: allocation}. 
By mapping the feasible set of motor velocity $\mathbb{V}$ using Equation \eqref{eq: mapping}, we obtain the feasible set of control inputs $\mathbb{U}$.
The algorithm first checks if the tilt control inputs $\tor_x, \tor_y$ stay within the feasible subset $\mathbb{B}$, which is obtained by projecting the feasible set $\mathbb{U}$ on $\thrust=0, \tor_y=0$ using Fourier-Motzkin elimination\cite{Ziegler1994LecturesOP}.
Specifically, $\tor_x, \tor_y$ should satisfy
\begin{align}
    \sqrt{2}\kthrust\ld(\vmin^{2}-\vmax^{2})\leqslant\tor_{x} \pm \tor_{y} \leqslant  \sqrt{2}\kthrust\ld(\vmax^{2}-\vmin^{2}).\label{eq: setB}
\end{align}
If \eqref{eq: setB} is not satisfied, a scale factor $\scale\in(0,1)$ is computed such that $\scale[\tor_x,\tor_y]\in\partial\mathbb{B}$, i.e.,  $\scale[\tor_x,\tor_y]$ lies on the boundary of $\mathbb{B}$ to preserve the magnitude of the original input as much as possible.
Then, we check if the scaled tilt input and the original yaw input, $[\scale\tor_x, \scale\tor_y, \tor_z]$, lie in the feasible set $\mathbb{A}$ obtained by projecting the feasible set $\mathbb{U}$ on $\thrust=0$:
% \begin{align}
%     \vmin^{2}-\vmax^{2} \leqslant \left\{ 
%     \begin{aligned}
%         \frac{\sqrt{2}}{2\kthrust\ld}\tor_{x} \pm \frac{\sqrt{2}}{2\kthrust\ld}\tor_{y} \\
%         \frac{\sqrt{2}}{2\kthrust\ld}\tor_{x} \pm \frac{1}{2\kdrag}\tor_{z} \\
%         \frac{\sqrt{2}}{2\kthrust\ld}\tor_{y} \pm \frac{1}{2\kdrag}\tor_{z} 
%     \end{aligned}
%     \right \} \leqslant  \vmax^{2}-\vmin^{2}\label{eq: setA}
% \end{align}
\begin{align}
    \vmin^{2}-\vmax^{2} \leqslant \left\{ 
    \begin{aligned}
        &\frac{\sqrt{2}\scale}{2\kthrust\ld}\tor_{x} \pm \frac{1}{2\kdrag}\tor_{z} \\
        &\frac{\sqrt{2}\scale}{2\kthrust\ld}\tor_{y} \pm \frac{1}{2\kdrag}\tor_{z} 
    \end{aligned}
    \right \} \leqslant  \vmax^{2}-\vmin^{2}.\label{eq: setA}
\end{align}
Similarly, if \eqref{eq: setA} is not satisfied, we scale $\tor_z$ with a factor $\beta$ such that $[\scale\tor_x, \scale\tor_y, \beta{\tor}_z]\in\partial\mathbb{A}$. 
Next, we compute the feasible range for $\thrust$ given the torque inputs $\scale\tor_x, \scale\tor_y, \beta{\tor}_z$:

\begin{align}
    \begin{aligned}
    &\thrust \geq 4\kthrust\vmin^{2} -\min\left\{ 
    -\frac{\sqrt{2}}{\ld}\alpha|\tor_{x}| \pm \frac{\sqrt{2}}{\ld}\alpha\tor_{y} \pm \frac{\kthrust}{\kdrag}\beta\tor_{z}
    \right \} \\
    &\thrust \leq 4\kthrust\vmax^{2} -\min\left\{ 
    -\frac{\sqrt{2}}{\ld}\alpha|\tor_{x}| \pm \frac{\sqrt{2}}{\ld}\alpha\tor_{y} \pm \frac{\kthrust}{\kdrag}\beta\tor_{z}
    \right \}. 
    \end{aligned}
    \label{eq: setU}
\end{align}
In theory, we can simply set $T$ to the lower bound of Equation \eqref{eq: setU} to minimize thrust input.
In practice, we add a small bias $\thrust_\text{ground}$ to $\thrust$ because the thrust coefficient $\kthrust$, obtained by fitting the full thrust curve, does not approximate the thrust well when the motors operate at a very low velocity, causing larger control error as will be shown in the experiments (Section \ref{subsec: attitude}). 
%If one of the above inequalities is violated, $\thrust$ is set to the value of the violated constraint.
Finally, the motor velocities are obtained from the inverse mapping of the modified control inputs, which are guaranteed to satisfy the velocity constraints.
\begin{algorithm}
\If{$[\tor_{x},\tor_{y}]^{T}\in \mathbb{B}$}
{
$\scale=1$
}
\Else
{
find $\scale\in(0,1)$ s.t. $\alpha[\tor_{x},\tor_{y}]^{\top} \in \partial \mathbb{B}$
}
\If{$[\scale\tor_{x},\scale\tor_{y},\tor_z]^{T}\in \mathbb{A}$}
{
$\beta=1$
}
\Else
{
find $\beta$ s.t. $[\alpha\tor_{x},\alpha\tor_{y}, \beta{\tor}_z]^{\top} \in \partial \mathbb{A}$
}
set $\thrust$ to the lower bound of Eqn. \eqref{eq: setU} \\
$\thrust = \thrust+\thrust_\text{ground}$  \tcp{add small bias}
% \If{$[\thrust,\scale\tor_{x},\scale\tor_{y},\beta\tor_z]^{T}\notin \mathbb{U}$}
% {
% find $\thrust$ s.t. $[\thrust, \alpha\tor_{x},\alpha\tor_{y}, \beta{\tor}_z]^{\top} \in \partial \mathbb{U}$
% }
% \If{$\tor \in \mathbb{A}$}
% {
% $\tor^{'} = \tor;$
% }
% \Else
% {
%     \If{$[\tor_{x},\tor_{y}]^{T}\notin \mathbb{B}$}
%     {
%         Find $\alpha$, $0<\alpha<1$, let $\alpha[\tor_{x},\tor_{y}]^{T} \in \partial \mathbb{B};$ \\
%         $[\tor_{x}^{'},\tor_{y}^{'}]^{T} = \alpha[\tor_{x},\tor_{y}]^{T};$\\
%         \If{$[\tor_{x}^{'},\tor_{y}^{'},\tor_{z}]\in \mathbb{A}$}{ $\tor_{z}^{'} = \tor_{z}$}
%         \Else{
%         Find $\tor_{z}^{'}$, let $\tor^{'}\in\mathbb{A}$
%         }
%     }
%     \Else
%     {
%         Find $\beta$, $0<\beta<1$,let $[\tor_{x},\tor_{y},\beta\tor_{z}]^{T} \in \partial \mathbb{A};$ \\
%         $\tor^{'} = [\tor_{x},\tor_{y},\beta\tor_{z}]^{T};$
%     }
% }
% $\min \thrust^{'}, \; s.t. \; \ctrlin^{'} =[\thrust^{'},\tor^{'}]^{T}\in \mathbb{U}; $ \\
% $\thrust^{'}=\thrust^{'}+\thrust_{bias};$\\
    $\motorvelovect = \vb{M}^{-1}[\thrust, \alpha\tor_{x},\alpha\tor_{y}, \beta{\tor}_z]^\top$ \tcp{inverse map}

\caption{Control allocation algorithm}\label{alg: allocation}
\end{algorithm}

%It's worth noting that we add a bias to the minimum value of thrust $\thrust^{'}$ and we can explain it as below. The $\ctrlin^{'}$ with minimized $\thrust^{'}$ must lies on the boundary of $\mathbb{U}$, $\ctrlin^{'} \in u^{*}$ and according to the mapping relation it is easy to get $v \in v^{*}$, witch means at least one motor's speed is at minimum in one control loop. It seems no problem as both thrust and torque generated by a propulsion group, in \eqref{eq: thrust} and \eqref{eq: drag} is linearly related to $\motorvelo$. However, according to the thrust curve of motors, the aerodynamics in the smaller speed range is poor, which is more like a quadratic relation. Thus, the actual control efforts generated may not fit the desired one while the speed of the motor is working in a range near minimum speed.  Hereby, a bias is added to the minimum thrust needed to achieve linear relation between motor speed and thrust, to generate control efforts in more rationally.

% \subsection{Air-Ground Transition}
\section{Experiments}\label{sec: exp}
\subsection{Attitude Control and Power Consumption} \label{subsec: attitude}

\begin{figure}[]
      \centering      \includegraphics[width=\linewidth]{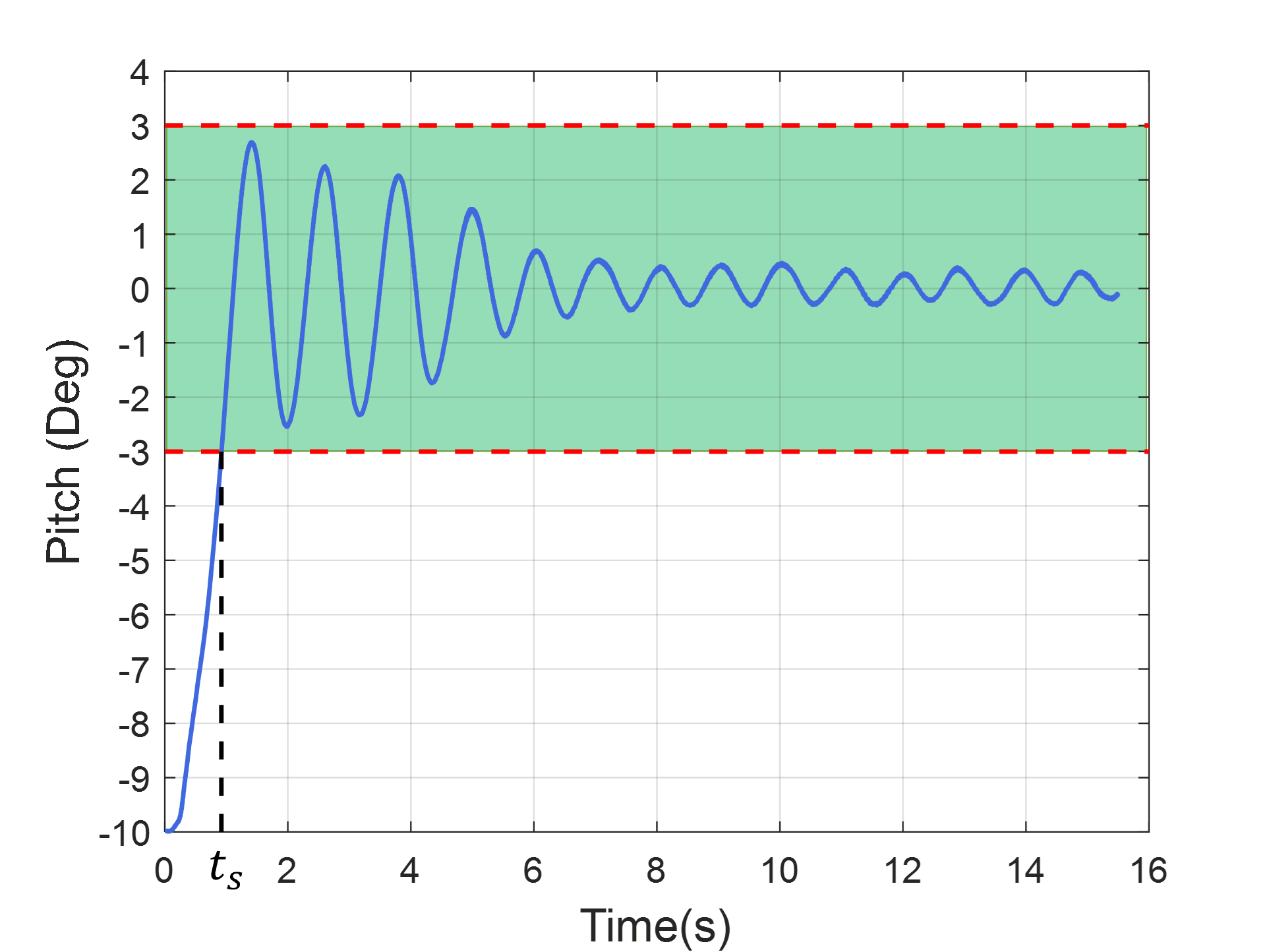}    
      \caption{\footnotesize The response of proposed method with ${\thrust_\text{ground}}/{\thrust_\text{max}} = 7.5\%$, the settling time $t_s = 0.9195s$. In this figure, $0\degree$ is the desired state and the region of $\pm3\degree$ error is colored in green.}
      \label{fig: settle_75}
\end{figure}

\begin{figure}[]
      \centering      \includegraphics[width=\linewidth]{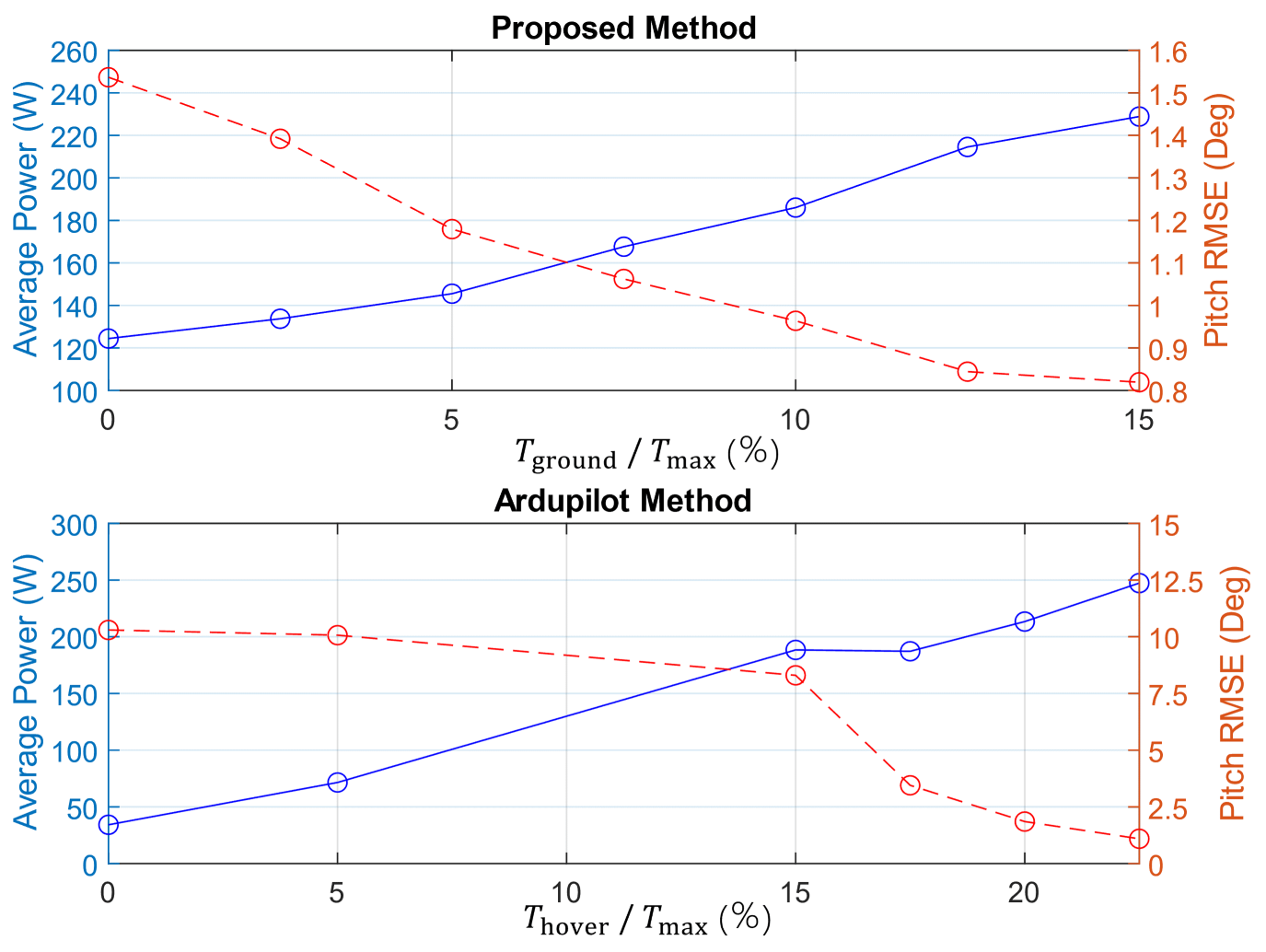}    
      \caption{\footnotesize Pitch angle error and power consumption under various throttle levels.  }
      \label{fig: pow_error_diff}
\end{figure}

We first conduct experiments to validate the attitude control performance of AirCrab in ground mode.
During the experiment, the robot starts at a rest position with a large pitch angle and tries to reach the level condition ($\roll,\pitch=0$).
%Multiple runs of experiments are conducted with varying values of $\thrust_\text{ground}$,
The attitude and power consumption of the robot are recorded with varying values of $\thrust_\text{ground}$.
For comparison, we also conduct the same experiment using the default attitude control algorithm of the widely used open-source Ardupilot firmware \footnote{https://ardupilot.org/copter/} and vary the hover throttle level (by adjusting the parameter MOT\_THST\_HOVER).
Figure \ref{fig: settle_75} shows the attitude of the robot for ${\thrust_\text{ground}}/{\thrust_\text{max}} = 7.5\%$.
The system takes $0.9195$ s to reach and stay within $\pm 3\degree$, and the pitch error continues to reduce and maintain within $\pm1\degree$ after $5.26$ s.
%The settling time is computed as the time taken for the attitude to stay within $\pm 3\degree$, 
The root mean square error (RMSE) of the pitch angle for $10$ s is computed for each experiment.
Figure \ref{fig: pow_error_diff} plots the trend of the power consumption and the RMSE for the proposed method and the Ardupilot method.
As we increase $\thrust_\text{ground}$, the pitch RMSE decreases while the power consumption increases for the proposed method. 
As explained in Section \ref{subsec: allocation}, the improvement in accuracy is due to a better approximation of the thrust using the experimentally determined thrust coefficient at higher motor velocity. 
In comparison, using the Ardupilot controller, the pitch angle could not stabilize within $\pm3\degree$ for a hovel throttle smaller than $17.5\%$.
This is because the control allocation prioritizes maintaining the altitude over attitude, which is reasonable for aerial operation.
However, when the thrust input is small, the control allocation scales down the rotation torque to maintain a small total thrust, causing insufficient torque to bring the robot to a level condition.  
When the hover throttle is above $17.5\%$, the pitch RMSE reduces but is still higher than the proposed method, and the energy consumption is higher.
Compared with the proposed method at ${\thrust_\text{ground}}/{\thrust_\text{max}} = 7.5\%$, 
the Ardupilot method produces $180\%,96.2\%$, and $12.1\%$ more error and consumes $11.7\%,23.5\%$, and $47.1\%$ more energy at hover throttle of $17.5\%, 20\%$, and $22.5\%$, respectively.
%At $22.5\%$ throttle level, the Ardupilot method produces $\%$ more error while consuming $\%$ more energy than the proposed method at ${\thrust_\text{ground}}/{\thrust_\text{max}} = 7.5\%$.
Hence, the proposed attitude controller and the control allocation method can achieve good control performance and high energy efficiency.
For subsequent experiments, we choose ${\thrust_\text{ground}}/{\thrust_\text{max}} = 7.5\%$ to exploit the good energy efficiency without much compromise on the control accuracy.
\begin{figure*}[]
      \centering      \includegraphics[width=\linewidth]{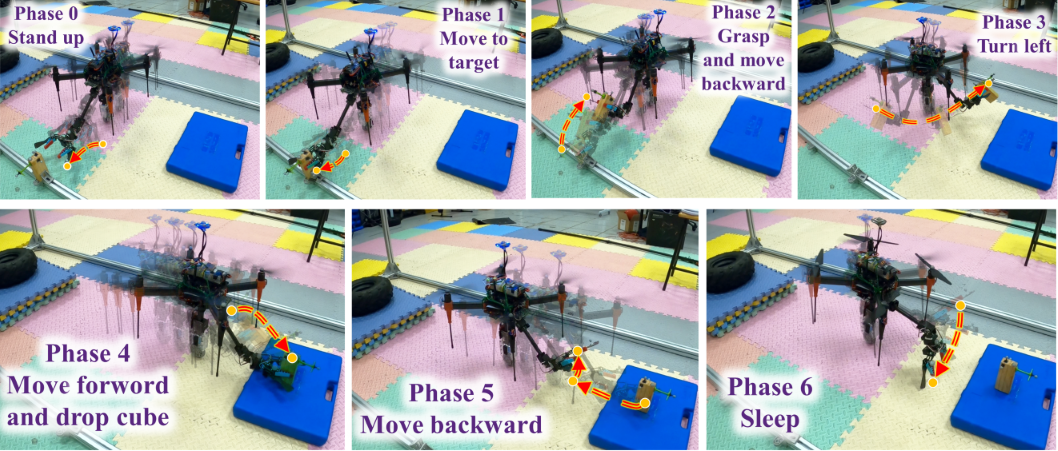}    
    \caption{\footnotesize The time-lapse photos of the manipulation experiment. The yellow points and red arrows illustrate the movement of the end effector.}
      \label{fig: ground_catching_photo}
\end{figure*}
\begin{figure}[]
      \centering      \includegraphics[width=\linewidth]{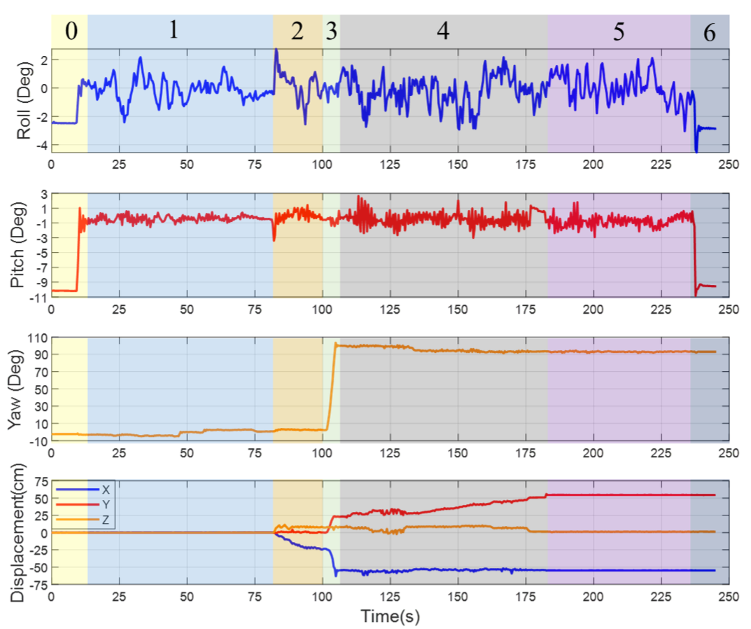}    
    \caption{\footnotesize The attitude of the AirCrab and the displacement of the object during the mission. Different phases of the mission are colored and numbered as Fig.\ref{fig: ground_catching_photo}.}
      \label{fig: ground_catching_curve}
\end{figure}

We then conduct an experiment in which the robot is manually controlled to move forward using the wheel, pick up an object of $90$ gram, move back and rotate $90\degree$, and then place the object.
Textured mats are used as uneven ground surfaces with bumps and ridges.
The attitude of the robot and the displacement of the object are recorded using the Vicon motion capture system and plotted in Figure \ref{fig: ground_catching_curve}.
The time-lapse images of the experiment are shown in Figure \ref{fig: ground_catching_photo}.
It is observed that the roll and pitch angles maintained within $\pm3\degree$ for most of the operation, and the yaw angle stayed within $3\degree$ error to the desired yaw.
A deviation of $3.9\degree$ is observed when the robot picks up the object, but the error reduces to less than $1\degree$ in $0.5$ s.
Oscillations of pitch angles are observed when the robot moves (phases $4$ and $5$) because the disturbance caused by the wheel rotation and uneven ground is not compensated in the controller.
The effect of the disturbance on roll and yaw is less significant as the wheel rotation overcomes friction forces mainly in the body $X$-axis.

\subsection{Manipulator Accuracy}
\begin{figure*}[]
      \centering      \includegraphics[width=\linewidth]{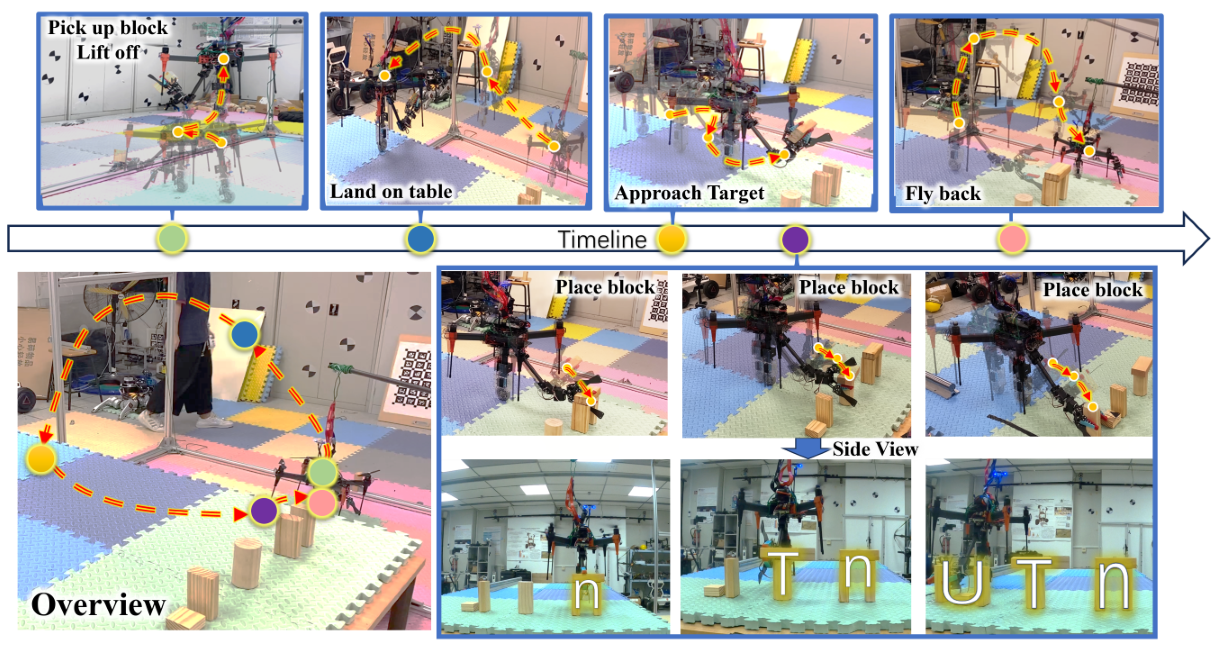}\caption{\footnotesize Overview of the hybrid aerial-ground manipulation experiments. In each experiment, one wooden is placed to complete one of the letters in `NTU'.}
      \label{fig: NTU}
\end{figure*}

\begin{figure}[]
      \centering      \includegraphics[width=\linewidth]{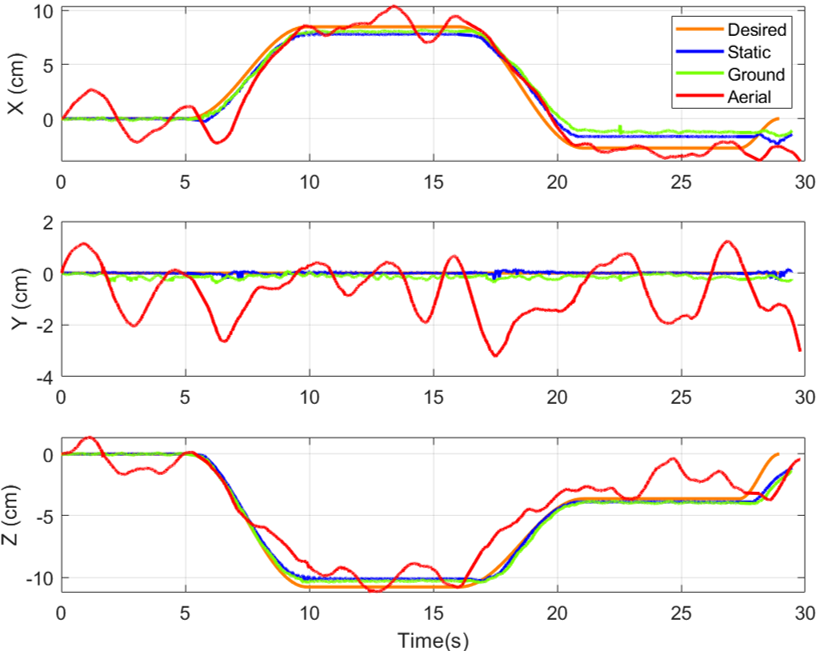}    
      \caption{\footnotesize Comparison of tracking performance of the end effector in different modes.}
      \label{fig: tracking}
\end{figure}

We measure the tracking accuracy of the end effector in three cases: (1) completely static, (2) ground mode, holding level attitude, and (3) aerial mode, holding position.
In aerial mode, the position estimate is obtained from an extended Kalman filter that fuses the onboard IMU and optical flow sensor.
We use onboard localization to simulate a realistic manipulation task where accurate external localization systems are unavailable.
In each case, the robot arm is commanded to execute a trajectory such that the end effector reaches $2$ setpoints sequentially and holds for $7$ seconds at each setpoint.
The trajectory setpoints are given in the robot arm's base frame (labeled with the subscript $a0$ in Fig. \ref{fig: axes}), and inverse kinematics is solved to obtain the target angles for each joint at $5$ Hz.
Figures \ref{fig: tracking} show the tracking performance for all modes.
%The trajectories are plotted in a static frame that coincides with the base frames of the robot arm at the time of starting the trajectory.
As expected, the tracking accuracy is best in static mode, with $0.90$ cm RMSE and $2.11$ cm max error. 
The tracking error in static mode arises due to the limitations of the robot arm, specifically, the low angular resolution of the low-cost servomotor joint and imperfect parameters used to compute kinematics.
The RMSE and maximum error in ground mode are $1.02$ cm and $2.14$ cm, respectively, slightly increased from those in static mode.
On the other hand, the RMSE and maximum error in aerial mode are significantly higher at $2.00$ cm and $5.53$ cm, respectively.
%The tracking RMSEs are $??\%$ and $??\%$ higher in ground mode and aerial mode than that in static mode, respectively.
From Figure \ref{fig: tracking}, it is observed that the tracking performance in ground mode is similar to that in the static mode.
Much larger position oscillation is observed in aerial mode, mainly due to drift in the position estimate and air disturbance. 
Therefore, ground contact is indeed beneficial in improving the manipulator's accuracy.
The high accuracy in ground mode enables challenging manipulation tasks as described in the next section.

\subsection{Hybrid Aerial-Ground Operation}

We conduct a series of hybrid aerial-ground manipulation tasks to demonstrate the capabilities of AirCrab. 
In each experiment, the robot is manually controlled to pick up a wooden block, take off and fly over a table, land, and place the block to complete one of the letters of the word NTU. 
A pilot remotely controls the ground and aerial locomotion, while another operator controls the joint angles of the arm. 
As a safety precaution, the robot is connected to a long rod through a safety tether, which does not exert force on the robot except during emergencies. 
Despite the manual operation, high accuracy and stability of the end effector are required to place the block in the designated location. Otherwise, the block may fall from the other blocks or fail to form the desired pattern. 
Figure \ref{fig: NTU} shows an overview of the tasks; a video of the experiments can be viewed in the supplementary document.

\section{CONCLUSIONS}\label{sec: conclusions}
In this work, we present the dynamics modeling and control design of AirCrab, a HAGM with a single active wheel.
Experiments have demonstrated the attitude control performance and the improvement in manipulation accuracy when using the proposed control strategy on the ground.
This work serves as a starting point for our subsequent research.
First, the control performance of the robot on various types of terrain could be enhanced by estimating the terrain slope and roughness.
Second, active disturbance rejection techniques could improve control accuracy under moving wheels and changing payload.
Eventually, we would like to realize a fully autonomous hybrid aerial-ground manipulation mission with onboard sensors and computational resources. 

% \addtolength{\textheight}{-12cm}   % This command serves to balance the column lengths
                                  % on the last page of the document manually. It shortens
                                  % the textheight of the last page by a suitable amount.
                                  % This command does not take effect until the next page
                                  % so it should come on the page before the last. Make
                                  % sure that you do not shorten the textheight too much.

%%%%%%%%%%%%%%%%%%%%%%%%%%%%%%%%%%%%%%%%%%%%%%%%%%%%%%%%%%%%%%%%%%%%%%%%%%%%%%%%

%%%%%%%%%%%%%%%%%%%%%%%%%%%%%%%%%%%%%%%%%%%%%%%%%%%%%%%%%%%%%%%%%%%%%%%%%%%%%%%%

%%%%%%%%%%%%%%%%%%%%%%%%%%%%%%%%%%%%%%%%%%%%%%%%%%%%%%%%%%%%%%%%%%%%%%%%%%%%%%%%
% \section*{APPENDIX}

% Appendixes should appear before the acknowledgment.

% \section*{ACKNOWLEDGMENT}

% The preferred spelling of the word ÒacknowledgmentÓ in America is without an ÒeÓ after the ÒgÓ. Avoid the stilted expression, ÒOne of us (R. B. G.) thanks . . .Ó  Instead, try ÒR. B. G. thanksÓ. Put sponsor acknowledgments in the unnumbered footnote on the first page.

\bibliographystyle{ieeetr}

\bibliography{reference}

\end{document}